\documentclass{article}
\usepackage{spconf,amsmath,graphicx}

\usepackage{tikz}

\usepackage{multirow}
\usepackage{booktabs}

\usepackage{xurl} 
\usepackage{url}  

\usepackage{enumitem}
\setlist{nosep, leftmargin=14pt}

\usepackage{mwe} 
\usepackage{xcolor} 

\title{Learning with Geometric Priors in U-Net Variants for \\Polyp Segmentation}
\vspace{-2mm}
\name{%
  \shortstack{
    Fabian Vazquez$^{1,\star}$, Jose A.~Nu\~nez$^{1,\star}$, Diego Adame$^{1}$, Alissen Moreno$^{1}$,
    Augustin Zhan$^{2}$, \\ Huimin Li$^{1}$, Jinghao Yang$^{1}$, Haoteng Tang$^{1,*}$, Bin Fu$^{1}$, Pengfei Gu$^{1,*}$  \thanks{$\star$ indicates equal contribution and $*$ indicates corresponding author}
    }
}
\vspace{-2mm}
\address{
  $^{1}$The University of Texas Rio Grande Valley, Edinburg, TX, USA \\
  $^{2}$Sewickley Academy, Sewickley, PA, USA
}
\begin{document}
%
\maketitle
\begin{abstract}
Accurate and robust polyp segmentation is essential for early colorectal cancer detection and for computer-aided diagnosis. While convolutional neural network-, Transformer-, and Mamba-based U-Net variants have achieved strong performance, they still struggle to capture geometric and structural cues, especially in low-contrast or cluttered colonoscopy scenes.
To address this challenge, we propose a novel Geometric Prior-guided Module (GPM) that injects explicit geometric priors into U-Net-based architectures for polyp segmentation. Specifically, we fine-tune the Visual Geometry Grounded Transformer (VGGT) on a simulated ColonDepth dataset to estimate depth maps of polyp images tailored to the endoscopic domain. These depth maps are then processed by GPM to encode geometric priors into the encoder’s feature maps, where they are further refined using spatial and channel attention mechanisms that emphasize both local spatial and global channel information.
GPM is plug-and-play and can be seamlessly integrated into diverse U-Net variants. Extensive experiments on five public polyp segmentation datasets demonstrate consistent gains over three strong baselines. {Code and the generated depth maps are available at: \texttt{https://github.com/fvazqu/GPM-PolypSeg}}
\end{abstract}
%

\begin{figure*}[t]
\centering
\includegraphics[width=0.98\textwidth]{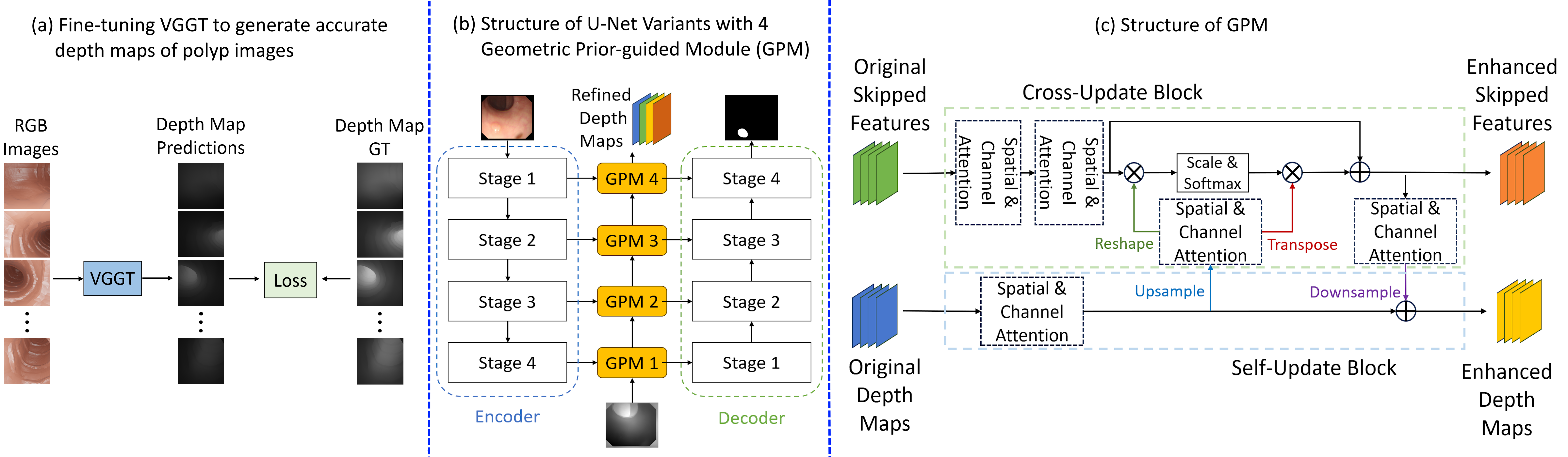}
\vspace{-0.35cm}
\caption{The pipeline of our proposed framework. (a) Visual Geometry Grounded Transformer (VGGT)~\cite{wang2025vggt} is fine tuned on a simulated endoscopic image dataset, the ColonDepth dataset~\cite{rau2019implicit} to generate high-fidelity, domain-adapted depth maps for polyp images. (b) Structure of U-Net variants with 4 Geometric Prior-guided Modules (GPMs) to refine skip connection features. (c) The structure of the GPM module.}
\label{fig:pipeline}
\end{figure*}
\vspace{-2mm}

%
\section{Introduction}
\label{sec:intro}
\vspace{-2mm}
Accurate medical image segmentation is essential for computer-aided diagnosis and treatment planning, particularly for the early prevention of colorectal cancer, one of the leading causes of cancer-related deaths worldwide~\cite{nunez2025white,vazquez2025exploring}. Automated polyp segmentation plays a crucial role by enabling precise localization and delineation of polyps, thereby supporting timely and accurate clinical interventions~\cite{fan2020pranet,adame2025topo}.

Deep learning (DL) has significantly advanced polyp segmentation. Convolutional neural network (CNN)-based architectures such as U-Net~\cite{ronneberger2015u} and its variants (e.g.,~\cite{fan2020pranet,gu2024self}) have demonstrated strong performance in medical imaging. However, CNNs are inherently limited by their local receptive fields, which restrict their ability to model long-range dependencies. Transformer-based models (e.g., U-Net v2~\cite{peng2025u}) address this limitation by capturing global context but often require large-scale datasets and substantial computational resources.
Recently, Vision Mamba-based networks have emerged as a new paradigm for efficient long-sequence modeling. The Mamba architecture~\cite{gu2023mamba} employs a linear-time state-space mechanism that captures both local and global dependencies with reduced computational cost. Building upon this idea, VM-UNetV2~\cite{zhang2024vm} introduced adaptive fusion layers and a hierarchical encoder–decoder design with stacked Vision Mamba (VM) blocks to enhance spatial awareness and multi-scale feature learning. Despite these advances, current models still struggle to capture geometric and structural cues, particularly in low-contrast or ambiguous polyp boundaries.

To address this limitation, we propose a novel plug-and-play Geometric Prior-guided Module (GPM) that injects explicit geometric priors into U-Net-based architectures for polyp segmentation.
This work tackles two pivotal questions:
\textbf{(I) How can we accurately generate endoscopy-tailored geometric priors that capture reliable scene geometry?
(II) How can these geometric priors be effectively encoded into U-Net-based architectures?
}

For question (I), we leverage the Visual Geometry Grounded Transformer (VGGT)~\cite{wang2025vggt} to generate high-fidelity, domain-adapted depth maps. Specifically, we fine-tune VGGT on a simulated endoscopic image dataset, the ColonDepth dataset~\cite{rau2019implicit}, to minimize the domain gap between nature scene and endoscopic images. The fine-tuned VGGT then produces depth maps for polyp images, providing robust geometric cues that enhance spatial reasoning and boundary precision.

For question (II), we design the GPM to incorporate these depth maps into the hierarchical feature maps generated by the encoder. GPM consists of two submodules: the Cross-Update Block (CUB) and the Self-Update Block (SUB). The CUB injects geometric priors (i.e., depth features) into the skip connections features by computing similarity-based weights between the features and depth maps. The SUB uses spatial- and channel-attention mechanisms~\cite{woo2018cbam} to emphasize local spatial structure and global channel context, refining the original skip connection features.

Our GPM can be seamlessly integrated into various U-Net variants to refine skip connection features. We evaluate the method on 5 public polyp segmentation datasets, and experimental results demonstrate that GPM consistently improves performance across three representative U-Net baselines.

\vspace{-2mm}
\section{Methods} 
\label{sec:methods}
\vspace{-2mm}
Fig.~\ref{fig:pipeline} shows the pipeline of our plug-and-play Geometric Prior-guided Module (GPM), which consists of two steps. First, we fine-tune VGGT~\cite{wang2025vggt} to produce high-fidelity, domain-adapted depth maps for polyp images. Second, GPM integrates these depth maps to inject geometric priors into the skip-connection features.

\vspace{-2mm}
\subsection{Depth Maps Generation}
\vspace{-2mm}
VGGT~\cite{wang2025vggt} is a large feed-forward transformer that directly predicts key 3D scene attributes—including camera intrinsics/extrinsics, depth maps, point maps, and 3D point tracks—from one to hundreds of input views in a single pass. It uses standard transformer blocks with minimal 3D inductive bias: images are patchified using DINOv2 features~\cite{oquab2023dinov2}, special camera tokens are appended, and the network alternates frame-wise and global self-attention. A lightweight camera head outputs camera parameters, while a DPT head~\cite{ranftl2021vision} produces dense per-frame predictions (depth/point maps and tracking features).

In this work, we employ VGGT to generate reliable geometric priors (depth maps). To narrow the gap between generic scenes and endoscopic imagery—characterized by low texture, specular highlights, smoke/water artifacts, and strong distortion—we fine-tune VGGT on simulated/annotated colon datasets (ColonDepth~\cite{rau2019implicit}). The fine-tuned model then yields endoscopy-tailored depth priors for polyp images (see Fig.~\ref{fig:pipeline}(a)). We choose VGGT over depth-only monocular models (e.g., Depth Anything~\cite{yang2024depth}) because it jointly predicts camera parameters and dense depth via a feed-forward transformer with a DPT head, providing stronger 3D consistency than depth-only backbones—crucial in endoscopy, where specularities, low texture, and repetitive patterns heighten scale/pose ambiguity. Its single- to multi-view flexibility also matches colonoscopy workflows (single images or short clips), and it fine-tunes cleanly on simulated endoscopic data to produce domain-adapted depth. These properties make VGGT a superior generator of endoscopy-tailored priors for our GPM compared with purely monocular alternatives.

\vspace{-2mm}
\subsection{Geometric Prior-guided Module (GPM)}
\vspace{-2mm}
As shown in Fig.~\ref{fig:pipeline}(c), the inputs to the GPM are the original skipped features $F_{o}$ and the geometric priors, i.e., the depth maps $D_{o}$. Through the Cross-Update Block and Self-Update Block, these inputs are refined into enhanced skipped features $F_{e}$ and enhanced geometric priors $D_{e}$, respectively. 

\textbf{Self-Update Block (SUB).} 
The SUB takes the raw depth maps $D_{o}$ and first applies a spatial- and channel-attention unit~\cite{woo2018cbam} to emphasize regions with significant depth discontinuities (e.g., polyp boundaries) and highlight the most informative geometric responses:
$
D'_{o} = \phi^c(\varphi^s(D_{o})),
$
where $\varphi^s$ and $\phi^c$ denote the spatial and channel attention parameters, respectively.
The refined depth $D'_{o}$ is then upsampled to match the resolution of the corresponding skip features, enabling cross-interaction in the CUB for geometry-guided feature refinement. In parallel, a second spatial- and channel-attention unit operates on a downsampled version of the refined skip features from the CUB to inject fine texture details into the refined depth map. Spatial attention enhances boundary contours and thin structures, while channel attention consolidates global context. The two depth representations are fused through element-wise addition:
$
D_{e} = D'_{o} + \phi^c(\varphi^s(Downsample(F_{e}))).
$
The resulting enhanced depth maps $D_{e}$ are propagated to the next stage as the input geometric priors, ensuring multi-scale geometric consistency across stages.

\textbf{Cross-Update Block (CUB).} 
The CUB first refines the skip-connection features $F_{o}$ by sequentially applying two spatial- and channel-attention units~\cite{woo2018cbam} to encode both local spatial structure and global channel context:
$
F'_{o} = \phi_2^c(\varphi_2^s(\phi_1^c(\varphi_1^s(F_{o})))).
$
The upsampled refined depth map $D'{o}$ from the SUB is also passed through a spatial- and channel-attention unit to produce geometry-aware embeddings, which are then used to compute a similarity map with the refined skip features:
$
C_{map} = Softmax\left(\frac{F'_{o} \times \phi^c(\varphi^s(Upsample(D'_{o})))}{\sqrt{N}}\right).
$
This similarity map is applied to the geometry-aware embedding to enhance the refined skip features with an identity connection:
$
F_{e} = C_{map} \times \phi^c(\varphi^s(Upsample(D'_{o}))) + F'_{o}.
$
The enhanced skip features $F_{e}$ are then passed to the encoder for decoding. Simultaneously, $F_{e}$ is forwarded to the SUB, where another spatial- and channel-attention unit uses it to refine the depth maps with high-frequency texture details.

Unlike the Shape Prior Module (SPM)~\cite{you2024learning}, which injects feature-derived shape priors (global and local) to regularize U-Net features, our GPM introduces externally estimated, domain-adapted geometric priors (depth) and fuses them bidirectionally with skip features. This design enforces stronger 3D geometry consistency and provides robustness against specularities, low texture, and ambiguous boundaries that are common in colonoscopy scenes.

\vspace{-2mm}
\subsection{Integrating GPMs into U-Net Variants}
\vspace{-2mm}
As illustrated in Fig.~\ref{fig:pipeline}(b), we integrate GPMs into U-Net variants by inserting one GPM at each encoder–decoder skip level (e.g., four GPMs for four stages), without modifying the backbone architecture or loss function. For each stage, the encoder’s skip features and either the VGGT-derived or previously enhanced depth maps are input to the GPM. The module outputs enhanced skip features that replace the original skip connections feeding the decoder at that level, and updated depth maps that are passed to the next GPM to maintain multi-scale geometric consistency. This plug-and-play integration is compatible with various U-Net variants, preserving the original topology, adding minimal parameters, and supporting end-to-end training. Consequently, the geometric priors sharpen object boundaries and stabilize predictions across scales. In this work, we integrate GPM into three representative polyp segmentation models: a CNN-based model (U-Net), a Transformer-based model (U-Net v2), and a Vision Mamba-based model (VM-UNetV2).

\vspace{-2mm}
\section{Experiments and Results} \label{exp}
\label{sec:exp}

\vspace{-2mm}
\subsection{Datasets}
\vspace{-2mm}
\textbf{VGGT Fine-tuning.}
We fine-tune VGGT on the ColonDepth dataset~\cite{rau2019implicit}, which contains 16{,}016 synthetic RGB–depth pairs simulating realistic colonoscopic views. All images are resized to $256\times256$, and depth values are min–max normalized to $[0,1]$. We use an 80/20 train–validation split.
We use the Scenario dataset~\cite{yang2023geometry} to evaluate the fine-tuned VGGT, which contains 4,500 RGB–depth pairs.

\noindent
\textbf{Polyp Segmentation.}
We conduct experiments on five public datasets: Kvasir-SEG~\cite{jha2020kvasir}, ClinicDB~\cite{bernal2015wm}, ColonDB~\cite{tajbakhsh2015automated}, ETIS~\cite{silva2014toward}, and CVC-300~\cite{vazquez2017benchmark}. For a fair comparison, we follow the train/test protocol in~\cite{peng2025u}. Specifically, the training set consists of 900 images from Kvasir-SEG and 550 images from ClinicDB. The test set includes all images from CVC-300 (60), ColonDB (380), and ETIS (196), Kvasir-SEG (100), and ClinicDB (62).

\vspace{-2mm}
\subsection{Experimental Setup}
\vspace{-2mm}
We fine-tune a pre-trained VGGT-1B model on colonoscopy images for depth prediction.
Images and depth maps are resized to a fixed resolution of $518\times518$.
Training is conducted on an NVIDIA A100 GPU (80\,GB) using AdamW (weight decay $=3\times10^{-3}$), with learning rate $1\times10^{-4}$, batch size $8$, and $100$ epochs; early stopping is used to mitigate overfitting.

All segmentation experiments are implemented in PyTorch and MONAI.
Models are trained on an NVIDIA A30 GPU (24\, GB) using AdamW (weight decay $=5\times10^{-3}$).
Following~\cite{zhang2024vm}, we resize all images to $256\times256$, set the learning rate to $1\times10^{-3}$, the batch size to $10$, and train for $350$ epochs.
We use \texttt{CosineAnnealingLR} with $T_{\max}=50$ and a minimum learning rate of $1\times10^{-5}$.
Standard data augmentation is applied to reduce overfitting.

For depth estimation, we report six metrics following~\cite{du2024polyp}: $\delta_1$, $\delta_2$, $\delta_3$ (accuracy), AbsRel, RMSE (mm), and $\log_{10}$ (mm) (error).
For polyp segmentation, we report Dice similarity coefficient (DSC) and intersection over union (IoU).
Each experiment is repeated five times with different random seeds, and we report the mean results.



\vspace{-2mm}
\subsection{Experimental Results}
\vspace{-2mm}
\textbf{Depth Map Generation.}
Table~\ref{tab:scenario_results} reports a quantitative comparison on the Scenario dataset.
The fine-tuned VGGT attains SOTA performance across all six evaluation metrics when comparing against the MDEM~\cite{du2024polyp}, GLPDepth~\cite{kim2022global}, and the original~VGGT~\cite{wang2025vggt}, indicating strong capability for producing reliable, endoscopy-tailored depth maps.

\noindent
\textbf{Polyp Segmentation.}
Table~\ref{tab:polyp} compares methods on five public polyp-segmentation datasets.
With GPM inserted, three representative backbones---a CNN-based model (U-Net), a Transformer-based model (U-Net v2), and a Mamba-based model (VM-UNetV2)---are \emph{consistently and significantly} improved, confirming the effectiveness of our GPM.
We do not include additional baselines (e.g., PraNet~\cite{fan2020pranet} etc.) since VM-UNetV2 represents a SOTA for polyp segmentation.

\begin{figure}[h!]
\centering
\includegraphics[width=0.45\textwidth]{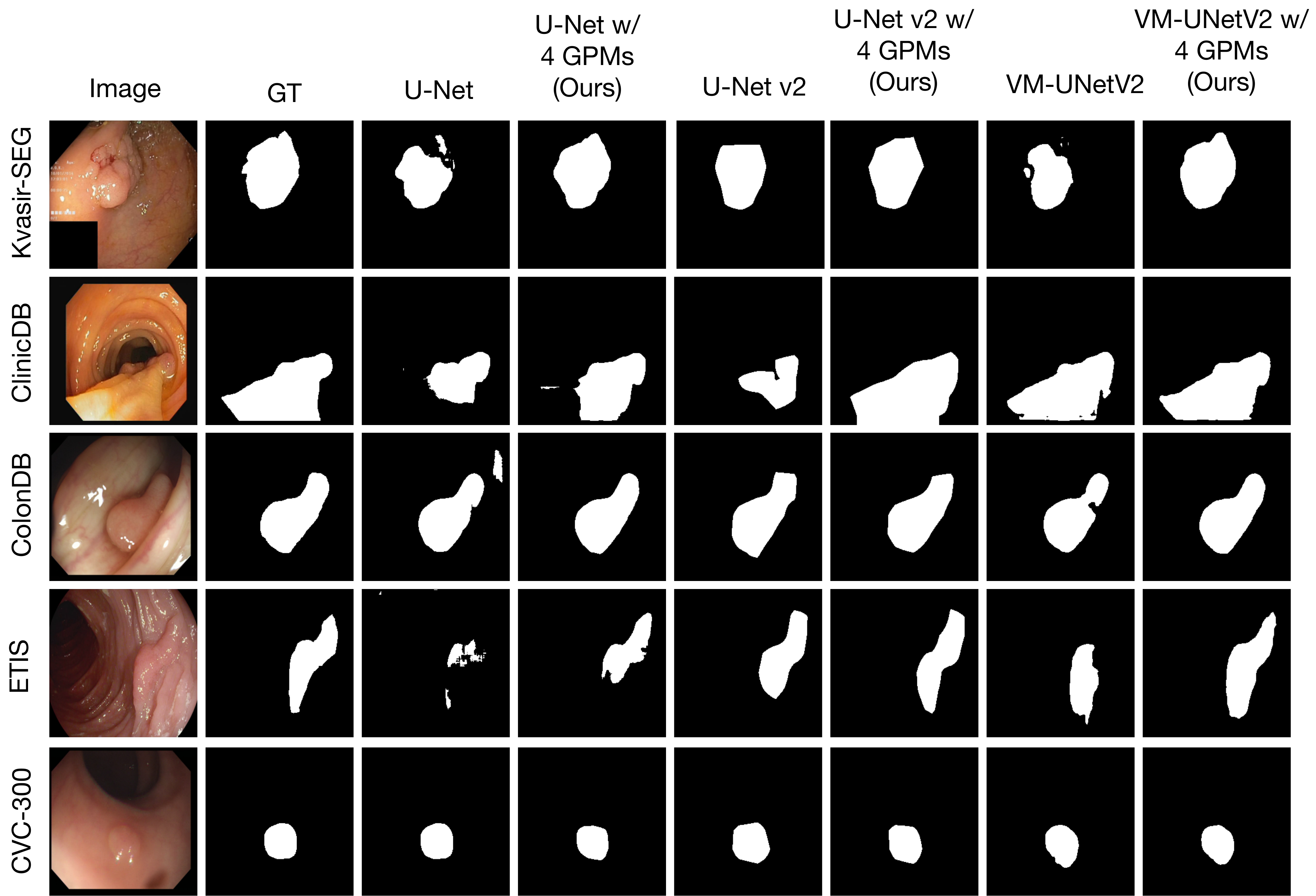}
\vspace{-0.35cm}
\caption{Visual examples of segmentations results.}
\label{fig:visual-results}
\end{figure}

\noindent
\textbf{Qualitative Results.}
Fig.~\ref{fig:visual-results} shows representative visualizations from all five datasets.
(1) Predictions with GPM align more closely with the ground truth than competing methods, and (2) boundary regions are notably sharper—especially under low contrast in ClinicDB and ETIS—demonstrating that encoding geometric priors into U\!-Net variants enhances boundary fidelity and overall segmentation quality.


\begin{table}[t]
  \centering
  \caption{{The quantitative depth map generation results for different methods. The best results are in bold.}}
  \label{tab:scenario_results}
\vspace{0.05cm}
   \scalebox{0.62}{
  \begin{tabular}{lcccccc}
    \toprule
    {Methods} &
    $\delta_{1}$ & $\delta_{2}$ & $\delta_{3}$ &
    $\mathrm{AbsRel}$ & $\log_{10}(\mathrm{mm})$ &
    $\mathrm{RMSE}(\mathrm{mm})$ \\
    \midrule
    MDEM~\cite{du2024polyp}   &0.446  &0.714   &0.862   &0.318   &0.143  &10.559  \\
   GLPDepth~\cite{kim2022global}   & 0.644 & 0.860  & 0.945  & 0.199  & 0.097 & 8.027 \\
    VGGT~\cite{wang2025vggt}   &0.734  &0.932   &\textbf{{0.977}}   &0.165   &0.075  &6.607  \\
    \addlinespace[2pt]
    Fine-tuned VGGT (Ours)       & \textbf{0.863} & \textbf{0.950} & {0.974} &
                  \textbf{0.123} & \textbf{0.058} & \textbf{3.812} \\
    \bottomrule
  \end{tabular}
  }
\end{table}
\begin{table}[t]
\centering
\caption{Experimental comparison of different methods on the five polyp segmentation datasets.}
\vspace{0.05cm}
 \scalebox{0.7}{
\begin{tabular}{l|llll}
\hline
Datasets & Methods & DSC (\%) $\uparrow$  & IoU (\%) $\uparrow$    \\
\hline 
\multirow{6}{*}{Kvasir-SEG} & 
U-Net~\cite{ronneberger2015u} &85.26	&77.79	  \\
& U-Net w/ 4 GPMs (Ours)&88.48 &81.90  
\\\cline{2-4}
& U-Net v2~\cite{peng2025u} &85.41	&77.36	 \\
& U-Net v2 w/ 4 GPMs (Ours)&88.14 &81.13
\\\cline{2-4}
& VM-UNetV2~\cite{zhang2024vm} &89.32	&82.85	  \\
& VM-UNetV2 w/ 4 GPMs (Ours)&\textbf{90.34} &\textbf{84.34}  \\
\hline 

\multirow{6}{*}{ClinicDB} & 
U-Net~\cite{ronneberger2015u} &84.08	&78.52	  \\
& U-Net w/ 4 GPMs (Ours)&87.38 &81.76  
\\\cline{2-4}
& U-Net v2~\cite{peng2025u} &83.59	&76.02	  \\
& U-Net v2 w/ 4 GPMs (Ours)&84.00 &77.09 
\\\cline{2-4}
& VM-UNetV2~\cite{zhang2024vm} &86.88	&81.15  \\
& VM-UNetV2 w/ 4 GPMs (Ours)&\textbf{88.23} &\textbf{82.68}  \\
\hline 

\multirow{6}{*}{ColonDB} & 
U-Net~\cite{ronneberger2015u} &65.67	&57.53	  \\
& U-Net w/ 4 GPMs (Ours)&72.04 &63.25 
\\\cline{2-4}
& U-Net v2~\cite{peng2025u} &68.71	&58.00	 \\
& U-Net v2 w/ 4 GPMs (Ours)&73.13 &63.70 
\\\cline{2-4}
& VM-UNetV2~\cite{zhang2024vm} &73.75	&64.63	  \\
& VM-UNetV2 w/ 4 GPMs (Ours)&\textbf{76.85} &\textbf{68.61} \\
\hline 

\multirow{6}{*}{ETIS} & 
U-Net~\cite{ronneberger2015u} &49.10	&42.21	  \\
& U-Net w/ 4 GPMs (Ours)&60.24 & 52.34 
\\\cline{2-4}
& U-Net v2~\cite{peng2025u} &55.85	&45.95  \\
& U-Net v2 w/ 4 GPMs (Ours)&62.25 &52.88 
\\\cline{2-4}
& VM-UNetV2~\cite{zhang2024vm} &64.49	&55.77	  \\
& VM-UNetV2 w/ 4 GPMs (Ours)&\textbf{70.10} &\textbf{61.62}  \\
\hline 

\multirow{6}{*}{CVC-300} & 
U-Net~\cite{ronneberger2015u} &77.93	&69.68	  \\
& U-Net w/ 4 GPMs (Ours)&78.51 &70.36 
\\\cline{2-4}
& U-Net v2~\cite{peng2025u} &79.25	&69.49	  \\
& U-Net v2 w/ 4 GPMs (Ours)&82.86 &74.71  
\\\cline{2-4}
& VM-UNetV2~\cite{zhang2024vm} &82.90	&74.57	  \\
& VM-UNetV2 w/ 4 GPMs (Ours)&\textbf{84.43} &\textbf{76.90}  \\\hline 
\end{tabular}
}
\label{tab:polyp}
\end{table}

\vspace{-2mm}
\subsection{Ablation Study}
\begin{figure}[h!]
\centering
\includegraphics[width=0.28\textwidth]{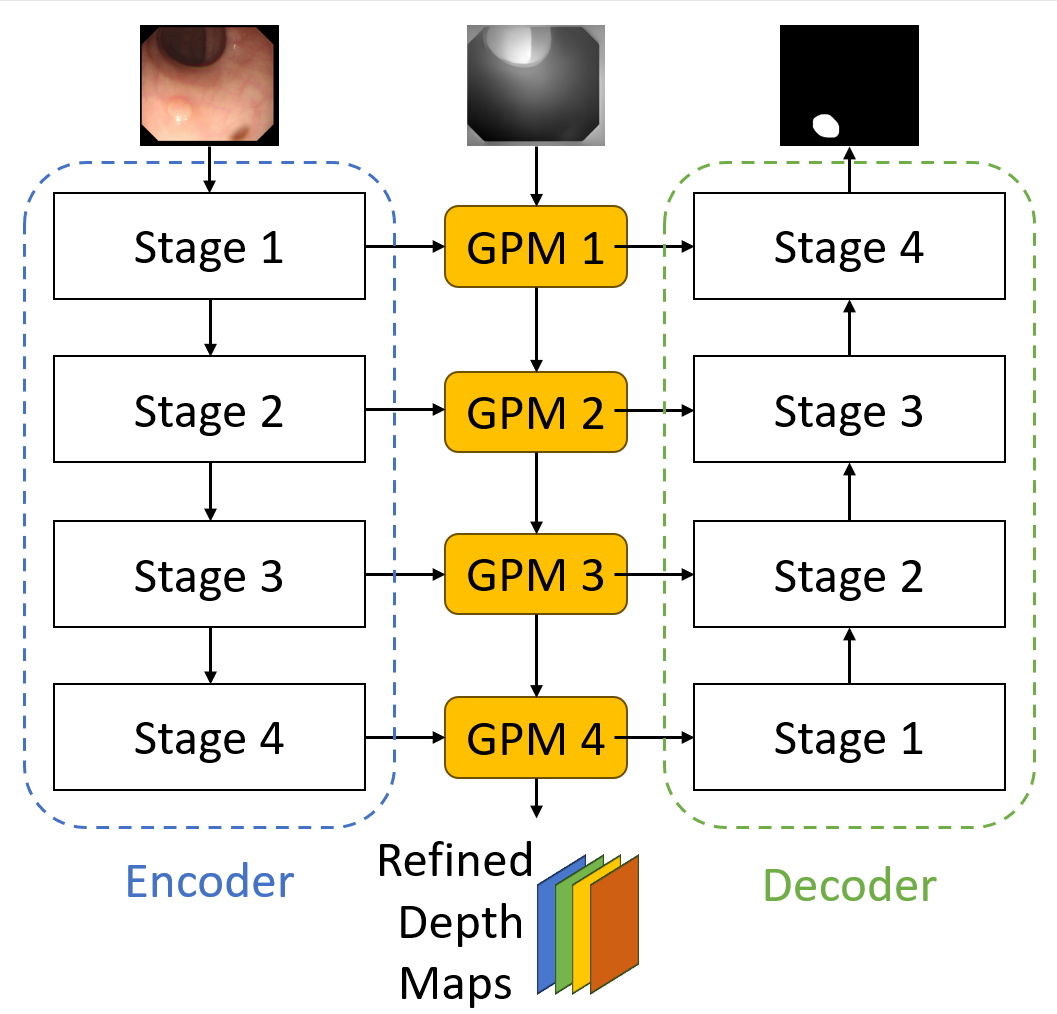}
\vspace{-0.35cm}
\caption{Architecture of U-Net Variant w/ 4 GPMs–Top.}
\label{fig:ablations}
\end{figure}

\vspace{-2mm}
We evaluate how GPM placement affects the encoding of geometric priors into skip features. 
Specifically, we compare two strategies (both using four GPMs across the four skip levels with VM-UNetV2~\cite{zhang2024vm}): 
(1) \emph{Top}\,$\rightarrow$\emph{Bottom} (shallowest to deepest skip; denoted {VM-UNetV2 w/ 4 GPMs–Top}, cf. Fig.~\ref{fig:ablations}), and 
(2) \emph{Bottom}\,$\rightarrow$\emph{Top} (deepest to shallowest; shown in Fig.~\ref{fig:pipeline}(b)). 
As summarized in Table~\ref{tab:ablation7col}, \emph{both} GPM-enhanced variants outperform the baseline VM-UNetV2, and the \emph{Bottom}\,$\rightarrow$\emph{Top} schedule (our default) yields a \emph{slight but consistent} gain over the \emph{Top}\,$\rightarrow$\emph{Bottom} variant.


\vspace{-0.35cm}
\begin{table}[t]
\centering
\caption{Ablation study with additional metrics.}
\vspace{0.05cm}
 \scalebox{0.54}{
\begin{tabular}{lccccccc}
\toprule
Method & Kvasir-SEG & ClinicDB & ColonDB & ETIS & CVC-300 & Average  \\
\midrule
VM-UNetV2~\cite{zhang2024vm} & 89.32 & 86.88 & 73.75 & 64.49 & 82.90 & 79.47  \\
VM-UNetV2 w/ 4 GPMs–Top  & \textbf{90.59} & 88.07 & \textbf{77.11} & 68.45 & \textbf{84.95} & 81.83 \\
VM-UNetV2 w/ 4 GPMs (Ours) & 90.34 & \textbf{88.23} & 76.85 & \textbf{70.10} & 84.43 & \textbf{81.99} \\
\bottomrule
\end{tabular}
}
\label{tab:ablation7col}
\end{table}

\vspace{-2mm}
\subsection{Computational Complexity}
\vspace{-2mm}
We report the number of parameters and FLOPs for all models.
As shown in Table~\ref{tab:metirc_model}, integrating GPM adds only a marginal overhead to U\!-Net (about 0.54M parameters). When GPM replaces the Semantics and Detail Infusion (SDI) module in U\!-Net v2 and VM-UNetV2, the total parameters slightly decrease, indicating that GPM is a lightweight alternative.


\begin{table}[]
\centering
\caption{Comparison of computational complexity.}
\label{tab:metirc_model}
\vspace{0.05cm}
 \scalebox{0.7}{
\begin{tabular}{cccc}
\hline
{Model} & {Input size} & {Params(M) $\downarrow$} & {FLOPs(G) $\downarrow$}  \\ \hline
U-Net          & (3, 256, 256)       &    \textbf{31.04}          & \textbf{48.23}               \\
U-Net w/ 4 GPMs (Ours)      & (3, 256, 256)       &  31.58          & 48.89                 \\\hline
U-Net v2         & (3, 256, 256)       &25.15               & \textbf{5.58}                \\
U-Net v2 w/ 4 GPMs (Ours)      & (3, 256, 256)       & \textbf{24.95}             &6.00                     \\\hline
VM-UNetV2      & (3, 256, 256)       &22.77      &\textbf{5.31}              \\ 
VM-UNetV2   w/ 4 GPMs (Ours)   & (3, 256, 256)       &\textbf{22.35}      &5.47      \\\hline
\end{tabular}
}
\end{table}

\vspace{-2mm}
\section{Conclusions} \label{concl}
\vspace{-2mm}
In this paper, we introduced a plug-and-play Geometric Prior-guided Module (GPM) that injects geometric cues into U-Net-style polyp segmentation models.
We first fine-tuned VGGT on the simulated ColonDepth dataset to produce endoscopy-tailored depth maps that capture reliable scene geometry.
Then, we used GPM to fuse these depth-derived priors into encoder features and refine them with complementary spatial and channel attention, strengthening both local structure and global context.
Experiments on five public datasets showed consistent improvements over strong three U-Net baselines, validating the effectiveness and generality of our approach.

\vspace{-2mm}
\section{Compliance with ethical standards}
\label{sec:ethics}
\vspace{-2mm}
This research study was conducted retrospectively using human subject data made available in open access by 7 publicly available datasets. Ethical approval was not required as confirmed by the licenses attached with the open access datasets.

\vspace{-2mm}
\section{Acknowledgements}
\vspace{-2mm}
This research was supported in part by NSF grant CCF-2523787.
We also acknowledge the UTRGV High Performance Computing Resource.


\small
\bibliographystyle{IEEEbib}
\bibliography{strings,refs}

\end{document}